\documentclass[nohyperref]{article}

\usepackage{microtype}
\usepackage{graphicx}
\usepackage{caption}
\usepackage{subcaption}
\usepackage{booktabs} 
\usepackage{multicol}

\usepackage{hyperref}

\usepackage[accepted]{icml2022}

\usepackage{amsmath}
\usepackage{amssymb}
\usepackage{mathtools}
\usepackage{amsthm}
\usepackage{wrapfig}

\usepackage[capitalize,noabbrev]{cleveref}

\theoremstyle{plain}

\theoremstyle{definition}

\theoremstyle{remark}

\usepackage{MnSymbol}%
\usepackage{wasysym}%
\usepackage{multirow}
\usepackage{xspace}
\usepackage{enumitem}
\def\arrvline{\hfil\kern\arraycolsep\vline\kern-\arraycolsep\hfilneg}
\usepackage[cal=dutchcal,
 calscaled=1,
 scr=euler]{mathalfa}

\usepackage{dsfont}

\newcommand{\mat}[1]{\ensuremath{{\mathbf{\MakeUppercase{{#1}}}}}}

\renewcommand{\vec}[1]{\ensuremath{\mathbf{\MakeLowercase{{#1}}}}}

\newcommand{\cv}{\vec{c}}
\newcommand{\xv}{\vec{x}}
\newcommand{\yv}{\vec{y}}

\def\Dt{\mathrm{D}}

\def\Nt{\mathrm{N}}
\def\Gt{\mathrm{G}}
\def\Kt{\mathrm{K}}

\def\sR{{\mathbb{R}}}

\definecolor{mydarkblue}{rgb}{0,0.08,0.45}
\hypersetup{colorlinks,citecolor={mydarkblue},urlcolor={mydarkblue}, linkcolor={red}} 

\let\oldcitet=\citet
\let\oldcitep=\citep 
\renewcommand{\citet}[1]{\textcolor{mydarkblue}{\oldcitet{#1}}}
\renewcommand{\citep}[1]{\textcolor{mydarkblue}{\oldcitep{#1}}}

\usepackage[textsize=tiny]{todonotes}

\icmltitlerunning{Towards a Single CNN Architecture to Model Long Range Dependencies in $\Nt$D}

\begin{document}

\onecolumn
\icmltitle{Towards a General Purpose CNN for Long Range Dependencies in $\Nt$D}



\icmlsetsymbol{equal}{*}

\begin{icmlauthorlist}
\icmlauthor{David W. Romero}{equal,vu}
\icmlauthor{David M. Knigge}{equal,uva}
\icmlauthor{Albert Gu}{stf}
\icmlauthor{Erik J. Bekkers}{uva}
\icmlauthor{Efstratios Gavves}{uva}
\icmlauthor{Jakub M. Tomczak}{vu}
\icmlauthor{Mark Hoogendoorn}{vu}
\end{icmlauthorlist}

\icmlaffiliation{vu}{Vrije Universiteit Amsterdam, The Netherlands}
\icmlaffiliation{uva}{University of Amsterdam, The Netherlands}
\icmlaffiliation{stf}{Stanford University, CA, USA}

\icmlcorrespondingauthor{David W. Romero}{d.w.romeroguzman@vu.nl}
\icmlcorrespondingauthor{David M. Knigge}{d.m.knigge@uva.nl}

\icmlkeywords{Machine Learning, ICML}

\vskip 0.3in



\printAffiliationsAndNotice{\icmlEqualContribution} 

\begin{abstract}
The use of Convolutional Neural Networks (CNNs) is widespread in Deep Learning due to a range of desirable model properties which result in an efficient and effective machine learning framework. However, performant CNN\break architectures must be tailored to specific tasks in order to incorporate considerations such as the input length, resolution, and dimentionality. In this work, we overcome the need for problem-specific CNN architectures with our \textit{Continuous Convolutional Neural Network} (CCNN): a single CNN architecture equipped with continuous convolutional kernels that can be used for tasks on data of arbitrary resolution, dimensionality and length without structural changes. Continuous convolutional kernels model long range dependencies at every layer, and remove the need for downsampling layers and task-dependent depths needed in current CNN architectures. We show the generality of our approach by applying the same CCNN to a wide set of tasks on sequential ($1\Dt$) and visual data ($2\Dt$).\break Our CCNN performs competitively and often outperforms~the~current~state-of-the-art across~all~tasks~considered.\vspace{-2mm}
\end{abstract}
\section{Introduction}
\vspace{-1mm}
Convolutional Neural Networks \citep{lecun1998gradient} (CNNs) are a class of Deep Learning models widely used for machine learning applications. Their popularity stems from their high performance and efficiency which has led them to achieve state-of-the-art in several applications across sequential \citep{abdel2014convolutional, van2016wavenet}, visual \citep{krizhevsky2012imagenet, simonyan2014very} and high-dimensional data \citep{schutt2017schnet, wu2019pointconv}.\break Nevertheless, an important limitation of CNNs --and Neural Networks in general-- is that their architectures must be tailored towards particular applications in order to handle different data lengths, resolutions and dimensionalities. This, in turn, has led to an extensive number of task-specific CNN architectures \cite{oord2016wavenet, bai2018empirical, simonyan2014very, szegedy2015going, ronneberger2015u, he2016deep, qi2017pointnet, wu2019pointconv}.

Data can come at many different lengths, e.g., images can be 32x32 or 1024x1024 and audio can easily be 16000 per second. The problem with standard CNNs is that their convolutional kernels are \textit{local}, which requires a custom architecture for each length with carefully chosen strides and pooling layers to capture full context. In addition, many types of data are inherently continuous in nature and have the same semantic meaning at different resolutions, e.g., images can be captured at arbitrary resolutions and have identical semantic content, and audio can be arbitrarily sampled at $16$kHz or $44.1$kHz and still sound the same to human ears. Nevertheless, conventional CNNs are bound to resolution and cannot be used across resolutions due to the \textit{discrete nature} of their convolutional kernels. Both problems are further exacerbated when considering data of different dimensionality with the same CNN, e.g., sequential ($1\Dt$), visual ($2\Dt$) and high-dimensional data ($3\Dt$, $4\Dt$), as different dimensionalities operate at different characteristic lengths and resolutions, e.g., a second of audio easily has length $16000$  which strongly contrasts with the size of images in benchmark datasets  \cite{krizhevsky2009learning, deng2009imagenet}.

\textbf{Towards a general-purpose CNN architecture.} In this work, we aim to construct a single CNN architecture that can be used on data of arbitrary resolutions, lengths and dimensionalities. 
Standard CNNs require task-specific architectures due to the discrete nature of their convolutional kernels which binds the kernels to specific data resolutions and makes them ill-suited to model global context due to the large amount of parameters required to construct large discrete convolutional kernels. Consequently, in order to construct a general-purpose CNN architecture it is crucial to develop a resolution agnostic convolutional layer able to model long range dependencies in a parameter efficient manner.

\begin{figure}
    \centering
    \hfill
         \begin{subfigure}[b]{0.15\textwidth}
         \centering
         \includegraphics[width=\textwidth]{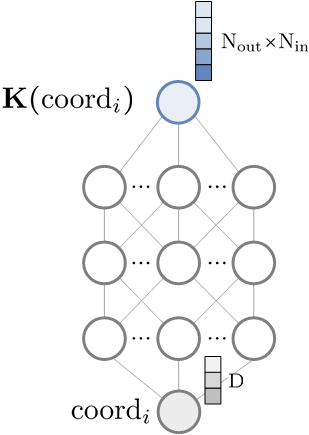}
         \caption{}
         \label{fig:ckconv_mlp}
     \end{subfigure}
     \hspace{4mm}
     \begin{subfigure}[b]{0.19\textwidth}
     \vspace{-5mm}
         \centering
         \includegraphics[width=\textwidth]{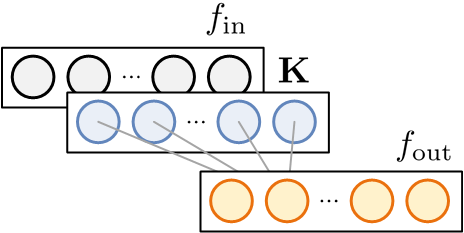}
         \caption{Sequential data}
         \label{fig:ckconv_1d}
     \end{subfigure}
     \hspace{1mm}
     \begin{subfigure}[b]{0.2\textwidth}
         \centering
         \includegraphics[width=\textwidth]{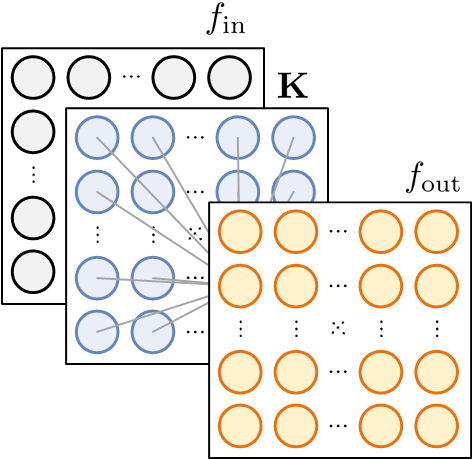}
         \caption{Visual data}
         \label{fig:ckconv_2d}
     \end{subfigure}
     \hspace{1mm}
     \begin{subfigure}[b]{0.2\textwidth}
         \centering
         \includegraphics[width=\textwidth]{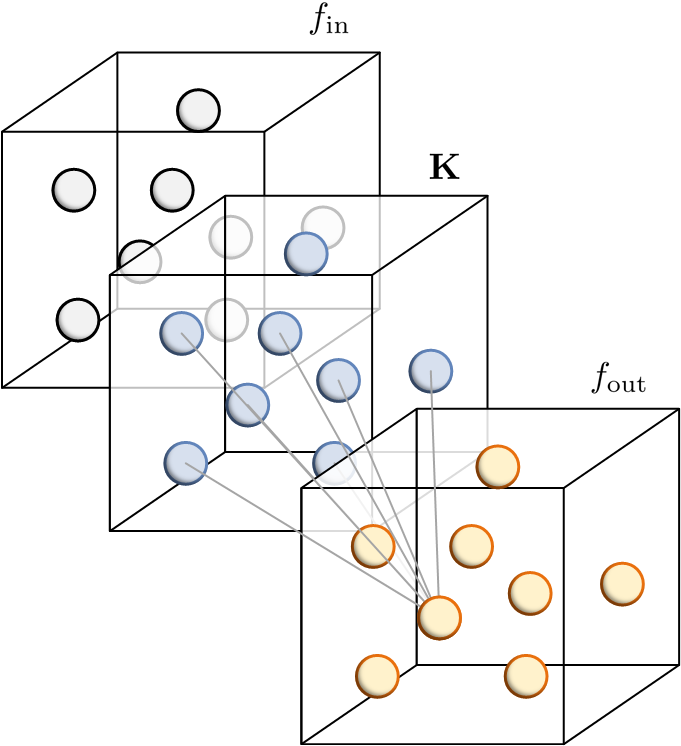}
         \caption{Point clouds}
         \label{fig:ckconv_3d}
     \end{subfigure}
    \hfill
    \vspace{-4mm}
    \caption{Continuous convolutional kernels. A continuous convolutional kernel is parameterized with a small neural network $\Gt_{\mathrm{Kernel}}$ that receives coordinates $\cv_i \in \sR^{\Dt}$ as input and outputs the value of the convolutional kernel at that position $\mat{K}(\cv_i) {=} \Gt_{\mathrm{Kernel}}(\cv_i) \in \sR^{\mathrm{N_{in}}\times\mathrm{N_{out}}}$ (\ref{fig:ckconv_mlp}). 
    The continuous parameterization of $\mat{K}$ allows the convolutional layer to (\emph{i)} model long range dependencies, (\emph{ii)} handle irregularly sampled data, and (\emph{iii)} be used across different resolutions. Additionally, changing the dimensionality of the coordinates $\cv_i$ can be used to construct convolutional kernels for sequential (\ref{fig:ckconv_1d}), visual (\ref{fig:ckconv_2d}), and higher dimensional data (\ref{fig:ckconv_3d}) with the same kernel generator network.
    \vspace{-4mm}}
    \label{fig:ckconv}
\end{figure}

\textbf{The need for continuous parameterizations.} Discrete convolutional kernels are defined with $\mathrm{N_{out}} {\times} \mathrm{N_{in}}$ independent learnable weights at each kernel position. Hence, large convolutional kernels require a large number of parameters and conventional CNNs rely on local convolutional kernels in combination with task-dependent depth values and pooling layers in order to model long range dependencies. 
Alternatively, we can construct \textit{continuous }convolutional kernels through use of a small neural network that maps positions to the value of the kernel at those positions (\citet{romero2022ckconv}, Fig.~\ref{fig:ckconv_mlp}). This approach decouples the size of the convolutional kernel from the number of parameters required to construct it, thus allowing the construction of arbitrary long kernels in a parameter efficient manner. Moreover, this parameterization overcomes the discrete nature of standard kernels and allows for the construction of resolution agnostic convolutional kernels that operate on coordinates of arbitrary resolution. 
Consequently, the same kernel generator network --and thus the same CNN-- can be used regardless of the input length and resolution. Furthermore, the same kernel generator network can be used to construct convolutional kernels for sequential $\Dt{=}1$ (Fig.~\ref{fig:ckconv_1d}), visual $\Dt{=}2$ (Fig.~\ref{fig:ckconv_2d}) and higher dimensional tasks $\Dt{\geq}3$ (Fig.~\ref{fig:ckconv_3d}) simply by changing the dimensionality of the input coordinates. In summary, the properties of \textit{Continuous Convolutional Kernels} allow for the construction of a single CNN architecture that can be used across data lengths, resolutions and dimensionalities. 

\vspace{-5mm}
\textbf{Contributions.} 
\vspace{-1mm}
\begin{itemize}[topsep=0pt, leftmargin=*]
\itemsep0em
    \item We present the \textit{Continuous CNN} (CCNN): a simple, general purpose CNN that can be used across data resolutions and dimensionalities without structural modifications. Our CCNN matches and often surpasses the state-of-the-art on several sequential (1D) and visual tasks (2D), as well as on tasks with irregularly-sampled data and test-time resolution changes.
    \item To this end, we provide several improvements for existing continuous CNN methods \cite{romero2022ckconv, romero2022flexconv} that allow them to match current state-of-the-art methods, e.g., S4 \cite{gu2022efficiently}. Our improvements include changes to the initiali-\break zation of the kernel generator networks, and modifications to the convolutional layers and the overall~structure~of~the~CNN.
\end{itemize}
\vspace{-2mm}
\section{Continuous Kernel Convolutions}
\vspace{-1mm}
Continuous kernel convolutions \cite{romero2022ckconv} parameterize convolutional kernels as continuous functions by using a small neural network $\mathrm{G_{Kernel}}: \sR^{\Dt} \rightarrow \sR^{\mathrm{N_{out}} \times \mathrm{N_{in}}}$ as a kernel generator network. This network maps a coordinate $\cv_i \in \sR^{\Dt}$ to the value of the convolutional kernel at that position:  $\mathrm{G_{Kernel}}(\cv_i) \in \sR^{\mathrm{N_{out}} {\times} \mathrm{N_{in}}}$ (Fig.~\ref{fig:ckconv_mlp}). 
By passing a vector of $\Kt$ coordinates $[\cv_i]_{i \in [1, ..., \Kt]}$ through $\mathrm{G_{Kernel}}$, a convolutional kernel $\mat{K}$ of equal size can be constructed, i.e., $\mat{K}{=}[\mathrm{G_{Kernel}}(\cv_i)]_{i \in [1, ..., \Kt]}$. Subsequently, a convolution operation takes place between an input signal $\xv: \sR^{\Dt} \rightarrow \sR^{\mathrm{N_{in}}}$ and the generated convolutional kernel $\mat{K}:\sR^{\Dt}\rightarrow \sR^{\mathrm{N_{out}}\times\mathrm{N_{in}}}$ to construct an output feature representation $\yv: \sR^{\Dt} \rightarrow \sR^{\mathrm{N_{out}}}$. That is: $\yv {=} \mathrm{ConvNd}( \mat{K}, \xv)$.

\vspace{-2mm}
\subsection{Properties}
\vspace{-1mm}
\textbf{A general operation for arbitrary data dimensionalities.} By changing the dimensionality $\Dt$ of the input coordinates $\cv_i$, the kernel generator network $\Gt_{\mathrm{Kernel}}$ can be used to construct convolutional kernels of arbitrary dimensionality. Consequently, the same operation can be used to process sequential $\Dt{=}1$, visual $\Dt{=}2$ and higher dimensional data $\Dt{\geq}3$.

\textbf{Parameter and computation efficient modelling of long range dependencies at every layer.} We can use the kernel generator network $\mathrm{G_{Kernel}}$ to construct convolutional kernels as big as the input signal in order to model long range dependencies at every layer, i.e., $\mat{K}{=}[\mathrm{G_{Kernel}}(\cv_i)]_{i \in [1, ..., \Kt]}$; $\Kt{=}\mathrm{len}(\xv)$. The number of parameters in $\Gt_{\mathrm{Kernel}}$ is independent from the length of the convolutional kernel, and thus kernels of arbitrary size can be constructed under a fixed parameter count. Convolutions with large convolutional kernels can be efficiently computed using the \textit{convolution theorem}, which states that a convolution in the time domain equals a pointwise product in the frequency domain.

\textbf{Irregularly-sampled data.} For some applications, the input $\xv$ may not be on a regular grid, e.g., medical data. Discrete convolutional kernels are ill-suited for such applications as their value is only known at some preset positions and not for arbitrary coordinates $\cv_i$. Contrarily, continuous kernels are defined everywhere and thus can handle irregular~data~natively.

\textbf{Equivalent responses across input resolutions.} If the input signal $\xv$ undergoes a resolution change, e.g., audio initially observed at $8$KHz is now observed at $16$KHz,
convolving with a discrete convolutional kernel would yield different responses, as the kernel would cover a different subset of the input at each resolution.
On the other hand, continuous kernels are resolution agnostic, and thus able to recognize an input regardless of its resolution. When presenting an input at a different resolution, e.g., higher resolution, it is sufficient to pass a finer grid of coordinates through the kernel generator network in order to construct the same kernel at the corresponding resolution. For a signal $\xv$ and a continuous convolutional kernel $\mat{K}$ sampled at resolutions $\mathrm{r^{(1)}}$ and $\mathrm{r^{(2)}}$, the convolution at both resolutions are approximately equal up to a factor proportional to the resolution change \cite{romero2022ckconv}:
\begin{equation}\label{eq:multires}
\setlength{\abovedisplayskip}{-2pt}
\setlength{\belowdisplayskip}{3pt}
    \mathrm{ConvNd}\left( \mat{K}_\mathrm{r^{(2)}}, \xv_\mathrm{r^{(2)}}\right) \approx \left(\mathrm{r^{(1)}}/\mathrm{r^{(2)}}\right)^{\Dt}\mathrm{ConvNd}\left( \mat{K}_\mathrm{r^{(1)}}, \xv_\mathrm{r^{(1)}}\right).
\end{equation}
\textbf{Learning of hyperparameters.} A promising property of continuous kernels is that they enable the learning of parameters that must otherwise be treated as hyperparameters in CNNs with discrete kernels. FlexConvs \cite{romero2022flexconv}, for instance, define their convolutional kernels as the product of a kernel generator network $\Gt_\mathrm{kernel}$ and a trimmed Gaussian mask $\mathrm{Gauss}_{\mu, \sigma}$ with learnable parameters $\mu, \sigma$, i.e., $\mat{K}(\cv_i){=}\Gt_\mathrm{kernel}(\cv_i) \cdot  \mathrm{Gauss}_{\mu, \sigma}(\cv_i)$. The Gaussian mask defines the size of the kernel and thus, by learning the mask, one can effectively learn the size of the convolutional kernel during training. In\break concurrent work we observe that a similar strategy can be used to additionally learn the depth and~width~of~neural~networks.
\vspace{-2mm}
\section{The Continuous Convolutional Neural Network: Modelling Long Range Dependencies in $\Nt$D}

\begin{wrapfigure}{r}{0.34\textwidth}
    \centering
    \vspace{-3mm}
    \includegraphics[width=0.32\textwidth]{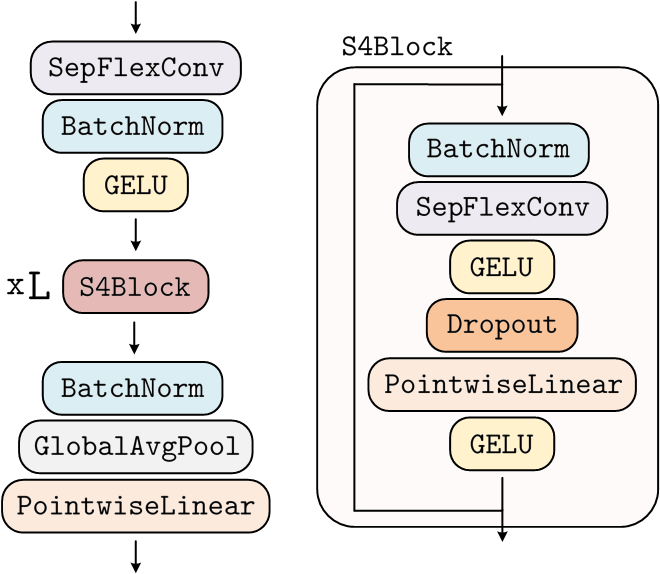}
    \vspace{-4mm}
    \caption{The Continuous CNN architecture.
    \vspace{-5mm}}
    \label{fig:ccnn}
\end{wrapfigure}
\vspace{-1mm}
\textbf{An improved residual block with continuous kernel convolutions.}
Recent works have shown that residual blocks \citep{he2016deep} can be strongly improved by changing the nonlinearities used and the position of the normalization layers within the blocks \cite{xiong2020layer, liu2022convnet}. Based on these observations, we modify the FlexNet architecture \citep{romero2022flexconv} with a residual network composed of blocks similar to those of S4 networks \citep{gu2022efficiently}. The CCNN architecture is shown in Figure ~\ref{fig:ccnn}.

\textbf{Depthwise separable continuous kernel convolutions.}
Separable convolutions have long been shown to improve the parameter and computational efficiency of CNNs \cite{rigamonti2013learning, sifre2014rigid}. 
More recently, their usage has shown improvement over conventional convolutions in CNNs \cite{chollet2017xception, tan2019efficientnet, knigge2021exploiting, liu2022convnet}, due to the separation of spatial and channel dimensions, which reduces the computational and parameter complexity of the convolution and allows for wider networks and higher performance.

Based on these observations we construct a depth-wise separable version of FlexConv \cite{romero2022flexconv}, in which a channel-wise convolution is computed with a kernel generated by a kernel generator network $\mathrm{G_{Kernel}}: \sR^{\Dt} \rightarrow \sR^{\mathrm{N_{in}}}$, followed by a point-wise convolution from $\mathrm{N_{in}}$ to $\mathrm{N_{out}}$. This change allows for the construction of much a wider CCNN --from 30 to 110 hidden channels-- without increasing the parameter or computation complexity of the network.

\textbf{Proper initialization of the kernel generator network $\mathrm{G_{Kernel}}$.}
We observe that the kernel generator networks in previous works are not properly initialized for their purpose of parameterizing convolutional kernels \cite{schutt2017schnet, wu2019pointconv}. Upon initialization, one would like the variance of the input and the output of a convolutional layer to remain equal to avoid exploding and vanishing gradients, i.e., $\mathrm{Var}(\xv){=}\mathrm{Var}(\yv)$. As such, convolutional kernels are initialized to have variance $\mathrm{Var}(\mat{K}){=}
\texttt{gain}^{2}/(\texttt{in\_channels} \cdot \texttt{kernel\_size})$, 
with a $\texttt{gain}$ that depends on the nonlinearity used \cite{he2015delving}. Nevertheless, neural networks are initialized such that the unitary variance of the input is preserved at the output. Consequently, when used as a kernel generator network, a standard initialization method leads the kernel to have unitary variance, i.e., $\mathrm{Var}(\mat{K}){=}1$. As a result, CNNs using neural networks as kernel generator networks experience a layer-wise growth in the variance of the feature representations proportional to ${\texttt{in\_channels} \cdot \texttt{kernel\_size}}$. For example, we observe that the logits of CKCNNs \cite{romero2022ckconv} and FlexNets \cite{romero2022flexconv} lie in the order of $1\mathrm{e}^{19}$ upon initialization. This is undesirable as it can lead to unstable training and the need for low learning rates.

To solve this problem, we require that the variance at the output of $\mathrm{G_{Kernel}}$ equals $\texttt{gain}^{2}/(\texttt{in\_channels} \cdot \texttt{kernel\_size})$ and not $1$. To this end and inspired by \citet{Chang2020Principled}, we re-weight the last layer of the kernel generator network by $\texttt{gain}/\sqrt{\texttt{in\_channels} \cdot \texttt{kernel\_size}}$. As a result, the variance at the output of the kernel generator network follows the initialization of conventional convolutional kernels, and the logits of CCNNs present unitary variance upon initialization.

\vspace{-2mm}
\section{Experiments and discussion}
\vspace{-1mm}
Our goal is to construct a single model that can be applied to data of arbitrary length, resolution and dimensionality. We construct two CCNNs of different sizes: CCNN$_{4, 110}$ (4 blocks, 110 channels) and CCNN$_{6, 380}$ (6 blocks, 380 channels) and validate them on several sequential ($1\Dt$) and visual ($2\Dt$) benchmark datasets. For sequential data, we consider 1D pixel-level image classification \cite{le2015simple, chang2017dilated}, speech classification \cite{warden2018speech} and the Long Range Arena (LRA) \cite{tay2021long} benchmark, which evaluates the capacity of models to describe long range dependencies. For visual data, we consider 2D classification of images of different size \cite{krizhevsky2009learning, coates2011analysis}. A complete description of the datasets used is given in Appx.~\ref{appx:dataset_description}. The hyperparameters used in our experiments as well as additional descriptions of our models are reported in Appx.~\ref{appx:empirical_details}.\footnote{Our code is publicly available at \href{https://github.com/david-knigge/ccnn}{\texttt{github.com/david-knigge/ccnn}}}  

\textbf{Results.} As shown in Tabs.~\ref{tab:1d_image_classif}-\ref{tab:lra_results}, our CCNN models perform well across all tasks considered. In fact, CCNNs set a new state of the art on multiple LRA tasks as well as 1D CIFAR10 pixel classification and raw speech classification on Speech Commands, while often being (much) smaller than competitive approaches.

\textbf{The importance of modelling long range dependencies on $\Nt$D.} In principle, we could consider all tasks as a sequential task in which no 2D structure is considered. This is done, for instance with S4 \citep{gu2022efficiently} due to the complexity of defining state spaces in multidimensional spaces. Nevertheless, this comes at the cost of throwing away important information regarding the nature of the data. Contrarily, CCNNs can be easily defined on multidimensional spaces simply by changing the dimensionality of the coordinates going into the kernel generator networks. Interestingly, we observe that by considering the 2D nature of the Image and Pathfinder tasks in the LRA benchmark, much better results can be obtained (Tab.~\ref{tab:2d_image_chassif}). In PathFinder with 2D images, our largest CCNN obtains an accuracy of $96.00$ outperforming the previous state of the art by a margin of almost 10\% points and performing remarkably better than the CCNN on flattened images. In addition, we observe that models trained on the original 2D data show faster convergence than their sequential counterparts (Fig.~\ref{fig:1dvs2d}).

We note that 2D CNNs with small convolutional kernels, e.g., ResNet-18, were unable to solve Pathfinder due to the lack of fine-grained global context modelling resulting from intermediate pooling layers. This was also seen by \citet{gu2020hippo}.

\begin{table}
\centering
\begin{minipage}{0.34 \textwidth}
\begin{center}
\caption{Pixel-level 1D image classifcation.}
\label{tab:1d_image_classif}
\vspace{-3.5mm}
\begin{small}
\scalebox{0.7}{
\begin{tabular}{cccccc}
\toprule
& \sc{Size}  & \sc{sMNIST} & \sc{pMNIST} & \sc{sCIFAR10} \\
 \midrule
  LSTM  &70\sc{k}& 87.2 & 85.7  & -\\
 GRU &70\sc{k}&   96.2 & 87.3  & - \\
 IndRNN  & 83\sc{k} & 99.0 & 96.0& -  \\
 DilRNN  & 44\sc{k} & 98.0 & 96.1 & - \\
 HiPPO-RNN & 500\sc{k} & - & 98.30 & - \\
  r-LSTM  & 500\sc{k} & 98.4 & 95.2 & 72.2 \\
 TCN &70\sc{k}& 99.0 & 97.2  & - \\
  TrellisNet & 8\sc{m} & 99.20 & 98.13 & 73.42\\
 Transformer &500\sc{k} & 98.9 & 97.9 & 62.2\\
LSSL & 7.8\sc{m} & 99.53 & \textbf{98.76} & 84.65 \\
S4 &  7.8\sc{m} & \textbf{99.63} & 98.70 & \textbf{91.13} \\
CKCNN & 98\sc{k} & 99.31 & 98.00 & 62.25 \\
CKCNN-Big & 1\sc{m} & 99.32 & 98.54 & 63.74 \\
FlexTCN-6 & 375\sc{k} & 99.62 & 98.63 & 80.82 \\
\midrule
CCNN$_{4, 110}$ & 200\sc{k} & \underline{\textbf{99.72}} & \textbf{98.82} &  \textbf{90.30}\\
CCNN$_{6, 380}$ & 2\sc{m} & \underline{\textbf{99.72}} & \underline{\textbf{98.84}} & \underline{\textbf{93.08}}  \\
\bottomrule
\end{tabular}}
\end{small}
\end{center}
\end{minipage}%
\hfill
\begin{minipage}{0.28\textwidth}
\begin{center}
\caption{Speech classification.}
\label{tab:speech_classif}
\vspace{-3.5mm}
\begin{small}
\scalebox{0.7}{
\begin{tabular}{cccccc}
\toprule
& \sc{Size}  & \sc{MFCC} & \sc{Raw} & 0.5x \\
 \midrule
Transformer &500\sc{k} & 90.75 & - & -\\
Performer & & 80.85 & 30.77 & 30.68  \\
ODE-RNN & & 65.90 & - & - \\
NRDE & & 89.80 & 16.49 & 15.12 \\
 ExpRNN & & 82.13 & 11.60 & 10.80 \\
 LipschitzRNN & & 88.38 & - & - \\
 LSSL & 300\sc{k} & 93.58 & - & - \\
 S4 & 300\sc{k} & 93.96 & \textbf{98.32} & \textbf{96.30}\\
 WaveGAN-D & 26.3\sc{m} & - & 96.25 & - \\
 CKConv & 105\sc{k} & 95.30 & 71.66 & 65.96\\
 FlexTCN-6  & 373\sc{k} &\textbf{ 97.67} & 91.73 & - \\
 \midrule
CCNN$_{4, 110}$ & 200\sc{k} & \textbf{95.01} & \textbf{98.34} & \textbf{96.22} \\
CCNN$_{6, 380}$ & 2\sc{m} & \underline{\textbf{97.98}} & \underline{\textbf{98.44}} & \underline{\textbf{96.44}}   \\
\bottomrule
\end{tabular}}
\end{small}
\end{center}
\end{minipage}
\hfill
\begin{minipage}{0.35 \textwidth}
\begin{center}
\caption{2D image classification.}
\label{tab:2d_image_chassif}
\vspace{-3.5mm}
\begin{small}
\scalebox{0.7}{
\begin{tabular}{cccccc}
\toprule
& \sc{Size}  & \sc{CIFAR10} & \sc{CIFAR100} & \sc{STL10}  \\
\midrule
ResNet-44 & 660\sc{k} & 92.90 & 71.15 & -\\
ResNet-18 & 11.2\sc{M} & \textbf{94.92} & \underline{\textbf{77.50}} & \textbf{81.04} \\
Parabolic CNN & 502\sc{k} & 88.5 & 64.8 & 77.0  \\
Hamiltonian CNN & 264\sc{k} & 89.3 & 64.9 & 78.3 \\
CKConv & 630\sc{k} & 86.8 & - & - \\
FlexNet-6 & 670\sc{k} & 92.2 & - & -\\

\midrule
CCNN$_{4, 110}$ & 200\sc{k}  & \textbf{92.78} & 66.86 & \textbf{81.80}\\
CCNN$_{6, 380}$ & 2\sc{m}  & \underline{\textbf{95.20}} & \textbf{73.16} & \underline{\textbf{83.00}} \\
\bottomrule
\end{tabular}}
\vspace{2mm}
\scalebox{0.7}{
\begin{tabular}{ccc}
\toprule
\multicolumn{3}{c}{\sc{2D Long Range Arena Tasks}} \\
\toprule
& \sc{2DImage} & \sc{2DPathfinder} \\
\midrule
CCNN$_{4, 110}$ & \textbf{89.48} & \textbf{94.80}\\
CCNN$_{6, 380}$ & \underline{\textbf{91.12}} & \underline{\textbf{96.00}}\\
\bottomrule
\end{tabular}}
\end{small}
\end{center}
\end{minipage}
\vspace{-7mm}
\end{table}

\begin{table}
\centering
\begin{minipage}{0.6 \textwidth}
\centering
\caption{Long Range Arena.}
\label{tab:lra_results}
\begin{small}
\scalebox{0.7}{
\begin{tabular}{ccccccccc}
\toprule
&  \sc{Size} & \sc{ListOps}  & \sc{Text} & \sc{Retrieval} & \sc{Image} & \sc{Pathfinder} & \sc{Path-X} & \sc{Avg.}\\
 \midrule
Transformer &0.5\sc{m} & 36.37 & 64.27 & 57.46 & 42.44 & 71.40 & - & 53.66 \\
Reformer & & 37.27 & 56.10 & 53.40 & 38.07 & 68.50 & - & 50.56\\
BigBird & & 36.05 & 64.02 & 59.29 & 40.83 & 74.87 & - & 57.17 \\
Linear Trans. & & 16.13 & 65.90 & 53.09 & 42.34 & 75.30 & - & 50.46 \\
Performer & & 18.01 & 65.40 & 53.82 & 42.77 & 77.05 & - & 51.18  \\
FNet & & 35.33 & 65.11 &59.61 &38.67 &77.80 &- &54.52\\
Nystromförmer & &37.15 &65.52 &79.56 &41.58 &70.94 &- &57.46  \\
Luna-256 & & 37.25 &64.57 &79.29 &47.38 &77.72 &- &59.37  \\
S4 & &\underline{\textbf{58.35}} &\textbf{76.02} &\underline{\textbf{87.09}} &\textbf{87.26} &\textbf{86.05} &\underline{\textbf{88.10}} &\underline{\textbf{80.48}} \\
\midrule
CCNN$_{4, 110}$ & 200\sc{k} & \textbf{44.85} & \textbf{83.59} & - & \textbf{87.62} & \textbf{91.36} & - & 76.86 \\
CCNN$_{6, 380}$ &  2\sc{m} & \textbf{43.60} &  \underline{\textbf{84.08}} & - & \underline{\textbf{88.90}} & \underline{\textbf{91.51}} & - & \textbf{77.02} \\
\bottomrule
\end{tabular}}
\end{small}
\end{minipage}%
\hfill
\begin{minipage}{0.39 \textwidth}
\begin{center}
    \includegraphics[width=1.05\textwidth]{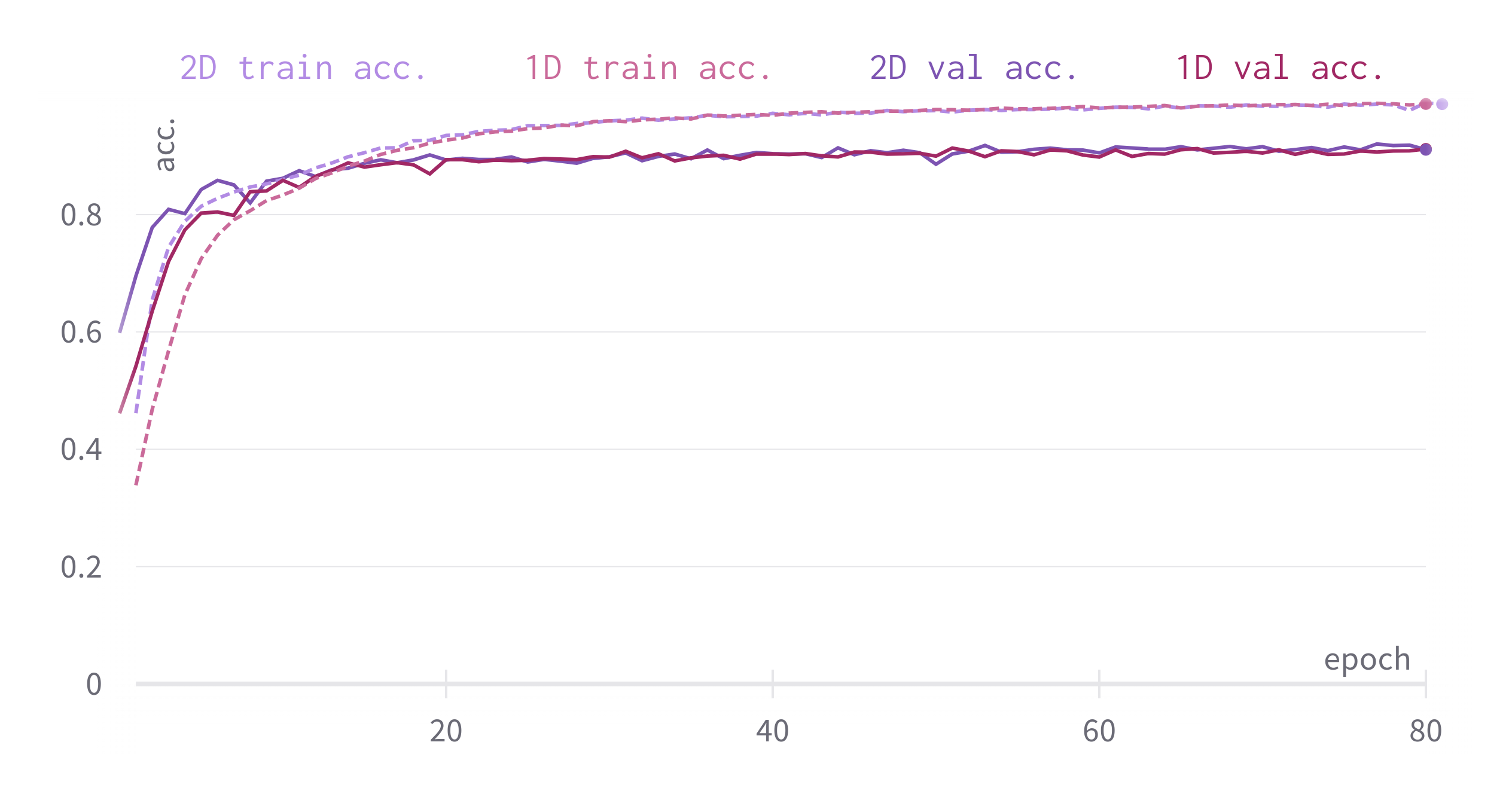}
    \vspace{-10mm}
    \captionof{figure}{1D vs. 2D image classification.}
    \label{fig:1dvs2d}
\end{center}
\end{minipage}
\vspace{-5mm}
\end{table}

\section{Conclusion}
\vspace{-1mm}
We propose the \textit{Continuous Convolutional Neural Network}: a single CNN architecture able to model long range dependencies on data of arbitrary length, resolution and dimensionality. Key to this development is the replacement of discrete convolutional kernels used in standard CNNs with Continuous Convolutional Kernels. With a single architecture, our CCNN performs competitively and often outperforms the current state of the art across a variety of sequential and visual tasks.
\vspace{-2mm}
\section*{Acknowledgements}
\vspace{-1mm}
This work is co-supported by the \href{https://www.qualcomm.com/research/university-relations/innovation-fellowship/2021-europe}{Qualcomm Innovation Fellowship} granted to David W. Romero. David W. Romero sincerely thanks Qualcomm for his support. David W. Romero is financed as part of the Efficient Deep Learning (EDL) programme (grant number P16-25), partly funded by the Dutch Research Council (NWO). David Knigge is partially funded by Elekta Oncology Systems AB and a RVO public-private partnership grant (PPS2102).

This work was carried out on the Dutch national infrastructure with the support of SURF Cooperative.


\bibliography{example_paper}

\begin{thebibliography}{41}
\providecommand{\natexlab}[1]{#1}
\providecommand{\url}[1]{\texttt{#1}}
\expandafter\ifx\csname urlstyle\endcsname\relax
  \providecommand{\doi}[1]{doi: #1}\else
  \providecommand{\doi}{doi: \begingroup \urlstyle{rm}\Url}\fi

\bibitem[Abdel-Hamid et~al.(2014)Abdel-Hamid, Mohamed, Jiang, Deng, Penn, and
  Yu]{abdel2014convolutional}
Abdel-Hamid, O., Mohamed, A.-r., Jiang, H., Deng, L., Penn, G., and Yu, D.
\newblock Convolutional neural networks for speech recognition.
\newblock \emph{IEEE/ACM Transactions on audio, speech, and language
  processing}, 22\penalty0 (10):\penalty0 1533--1545, 2014.

\bibitem[Bai et~al.(2018)Bai, Kolter, and Koltun]{bai2018empirical}
Bai, S., Kolter, J.~Z., and Koltun, V.
\newblock An empirical evaluation of generic convolutional and recurrent
  networks for sequence modeling.
\newblock \emph{arXiv preprint arXiv:1803.01271}, 2018.

\bibitem[Biewald(2020)]{wandb}
Biewald, L.
\newblock Experiment tracking with weights and biases, 2020.
\newblock URL \url{https://www.wandb.com/}.
\newblock Software available from wandb.com.

\bibitem[Chang et~al.(2020)Chang, Flokas, and Lipson]{Chang2020Principled}
Chang, O., Flokas, L., and Lipson, H.
\newblock Principled weight initialization for hypernetworks.
\newblock In \emph{International Conference on Learning Representations}, 2020.
\newblock URL \url{https://openreview.net/forum?id=H1lma24tPB}.

\bibitem[Chang et~al.(2017)Chang, Zhang, Han, Yu, Guo, Tan, Cui, Witbrock,
  Hasegawa-Johnson, and Huang]{chang2017dilated}
Chang, S., Zhang, Y., Han, W., Yu, M., Guo, X., Tan, W., Cui, X., Witbrock, M.,
  Hasegawa-Johnson, M.~A., and Huang, T.~S.
\newblock Dilated recurrent neural networks.
\newblock \emph{Advances in neural information processing systems}, 30, 2017.

\bibitem[Chollet(2017)]{chollet2017xception}
Chollet, F.
\newblock Xception: Deep learning with depthwise separable convolutions.
\newblock In \emph{Proceedings of the IEEE conference on computer vision and
  pattern recognition}, pp.\  1251--1258, 2017.

\bibitem[Coates et~al.(2011)Coates, Ng, and Lee]{coates2011analysis}
Coates, A., Ng, A., and Lee, H.
\newblock An analysis of single-layer networks in unsupervised feature
  learning.
\newblock In \emph{Proceedings of the fourteenth international conference on
  artificial intelligence and statistics}, pp.\  215--223. JMLR Workshop and
  Conference Proceedings, 2011.

\bibitem[Deng et~al.(2009)Deng, Dong, Socher, Li, Li, and
  Fei-Fei]{deng2009imagenet}
Deng, J., Dong, W., Socher, R., Li, L.-J., Li, K., and Fei-Fei, L.
\newblock Imagenet: A large-scale hierarchical image database.
\newblock In \emph{2009 IEEE conference on computer vision and pattern
  recognition}, pp.\  248--255. Ieee, 2009.

\bibitem[{Falcon et al.}(2019)]{falcon2019pytorch}
{Falcon et al.}, W.
\newblock Pytorch lightning.
\newblock \emph{GitHub. Note:
  https://github.com/PyTorchLightning/pytorch-lightning}, 3, 2019.

\bibitem[Fathony et~al.(2021)Fathony, Sahu, Willmott, and
  Kolter]{fathony2021multiplicative}
Fathony, R., Sahu, A.~K., Willmott, D., and Kolter, J.~Z.
\newblock Multiplicative filter networks.
\newblock In \emph{International Conference on Learning Representations}, 2021.
\newblock URL \url{https://openreview.net/forum?id=OmtmcPkkhT}.

\bibitem[Gu et~al.(2020)Gu, Dao, Ermon, Rudra, and R{\'e}]{gu2020hippo}
Gu, A., Dao, T., Ermon, S., Rudra, A., and R{\'e}, C.
\newblock Hippo: Recurrent memory with optimal polynomial projections.
\newblock \emph{Advances in Neural Information Processing Systems},
  33:\penalty0 1474--1487, 2020.

\bibitem[Gu et~al.(2022)Gu, Goel, and Re]{gu2022efficiently}
Gu, A., Goel, K., and Re, C.
\newblock Efficiently modeling long sequences with structured state spaces.
\newblock In \emph{International Conference on Learning Representations}, 2022.
\newblock URL \url{https://openreview.net/forum?id=uYLFoz1vlAC}.

\bibitem[He et~al.(2015)He, Zhang, Ren, and Sun]{he2015delving}
He, K., Zhang, X., Ren, S., and Sun, J.
\newblock Delving deep into rectifiers: Surpassing human-level performance on
  imagenet classification.
\newblock In \emph{Proceedings of the IEEE international conference on computer
  vision}, pp.\  1026--1034, 2015.

\bibitem[He et~al.(2016)He, Zhang, Ren, and Sun]{he2016deep}
He, K., Zhang, X., Ren, S., and Sun, J.
\newblock Deep residual learning for image recognition.
\newblock In \emph{Proceedings of the IEEE conference on computer vision and
  pattern recognition}, pp.\  770--778, 2016.

\bibitem[Kidger et~al.(2020)Kidger, Morrill, Foster, and
  Lyons]{kidger2020neural}
Kidger, P., Morrill, J., Foster, J., and Lyons, T.
\newblock Neural controlled differential equations for irregular time series.
\newblock \emph{Advances in Neural Information Processing Systems},
  33:\penalty0 6696--6707, 2020.

\bibitem[Knigge et~al.(2021)Knigge, Romero, and Bekkers]{knigge2021exploiting}
Knigge, D.~M., Romero, D.~W., and Bekkers, E.~J.
\newblock Exploiting redundancy: Separable group convolutional networks on lie
  groups.
\newblock \emph{arXiv preprint arXiv:2110.13059}, 2021.

\bibitem[Krizhevsky et~al.(2009)Krizhevsky, Hinton,
  et~al.]{krizhevsky2009learning}
Krizhevsky, A., Hinton, G., et~al.
\newblock Learning multiple layers of features from tiny images.
\newblock 2009.

\bibitem[Krizhevsky et~al.(2012)Krizhevsky, Sutskever, and
  Hinton]{krizhevsky2012imagenet}
Krizhevsky, A., Sutskever, I., and Hinton, G.~E.
\newblock Imagenet classification with deep convolutional neural networks.
\newblock \emph{Advances in neural information processing systems}, 25, 2012.

\bibitem[Le et~al.(2015)Le, Jaitly, and Hinton]{le2015simple}
Le, Q.~V., Jaitly, N., and Hinton, G.~E.
\newblock A simple way to initialize recurrent networks of rectified linear
  units.
\newblock \emph{arXiv preprint arXiv:1504.00941}, 2015.

\bibitem[LeCun et~al.(1998)LeCun, Bottou, Bengio, and
  Haffner]{lecun1998gradient}
LeCun, Y., Bottou, L., Bengio, Y., and Haffner, P.
\newblock Gradient-based learning applied to document recognition.
\newblock \emph{Proceedings of the IEEE}, 86\penalty0 (11):\penalty0
  2278--2324, 1998.

\bibitem[Liu et~al.(2022)Liu, Mao, Wu, Feichtenhofer, Darrell, and
  Xie]{liu2022convnet}
Liu, Z., Mao, H., Wu, C.-Y., Feichtenhofer, C., Darrell, T., and Xie, S.
\newblock A convnet for the 2020s.
\newblock \emph{arXiv preprint arXiv:2201.03545}, 2022.

\bibitem[Loshchilov \& Hutter(2016)Loshchilov and Hutter]{loshchilov2016sgdr}
Loshchilov, I. and Hutter, F.
\newblock Sgdr: Stochastic gradient descent with warm restarts.
\newblock \emph{arXiv preprint arXiv:1608.03983}, 2016.

\bibitem[Loshchilov \& Hutter(2017)Loshchilov and
  Hutter]{loshchilov2017decoupled}
Loshchilov, I. and Hutter, F.
\newblock Decoupled weight decay regularization.
\newblock \emph{arXiv preprint arXiv:1711.05101}, 2017.

\bibitem[Oord et~al.(2016)Oord, Dieleman, Zen, Simonyan, Vinyals, Graves,
  Kalchbrenner, Senior, and Kavukcuoglu]{oord2016wavenet}
Oord, A. v.~d., Dieleman, S., Zen, H., Simonyan, K., Vinyals, O., Graves, A.,
  Kalchbrenner, N., Senior, A., and Kavukcuoglu, K.
\newblock Wavenet: A generative model for raw audio.
\newblock \emph{arXiv preprint arXiv:1609.03499}, 2016.

\bibitem[Qi et~al.(2017)Qi, Su, Mo, and Guibas]{qi2017pointnet}
Qi, C.~R., Su, H., Mo, K., and Guibas, L.~J.
\newblock Pointnet: Deep learning on point sets for 3d classification and
  segmentation.
\newblock In \emph{Proceedings of the IEEE conference on computer vision and
  pattern recognition}, pp.\  652--660, 2017.

\bibitem[Rigamonti et~al.(2013)Rigamonti, Sironi, Lepetit, and
  Fua]{rigamonti2013learning}
Rigamonti, R., Sironi, A., Lepetit, V., and Fua, P.
\newblock Learning separable filters.
\newblock In \emph{Proceedings of the IEEE conference on computer vision and
  pattern recognition}, pp.\  2754--2761, 2013.

\bibitem[Romero et~al.(2022{\natexlab{a}})Romero, Bruintjes, Tomczak, Bekkers,
  Hoogendoorn, and van Gemert]{romero2022flexconv}
Romero, D.~W., Bruintjes, R.-J., Tomczak, J.~M., Bekkers, E.~J., Hoogendoorn,
  M., and van Gemert, J.
\newblock Flexconv: Continuous kernel convolutions with differentiable kernel
  sizes.
\newblock In \emph{International Conference on Learning Representations},
  2022{\natexlab{a}}.
\newblock URL \url{https://openreview.net/forum?id=3jooF27-0Wy}.

\bibitem[Romero et~al.(2022{\natexlab{b}})Romero, Kuzina, Bekkers, Tomczak, and
  Hoogendoorn]{romero2022ckconv}
Romero, D.~W., Kuzina, A., Bekkers, E.~J., Tomczak, J.~M., and Hoogendoorn, M.
\newblock {CKC}onv: Continuous kernel convolution for sequential data.
\newblock In \emph{International Conference on Learning Representations},
  2022{\natexlab{b}}.
\newblock URL \url{https://openreview.net/forum?id=8FhxBtXSl0}.

\bibitem[Ronneberger et~al.(2015)Ronneberger, Fischer, and
  Brox]{ronneberger2015u}
Ronneberger, O., Fischer, P., and Brox, T.
\newblock U-net: Convolutional networks for biomedical image segmentation.
\newblock In \emph{International Conference on Medical image computing and
  computer-assisted intervention}, pp.\  234--241. Springer, 2015.

\bibitem[Sch{\"u}tt et~al.(2017)Sch{\"u}tt, Kindermans, Sauceda~Felix, Chmiela,
  Tkatchenko, and M{\"u}ller]{schutt2017schnet}
Sch{\"u}tt, K., Kindermans, P.-J., Sauceda~Felix, H.~E., Chmiela, S.,
  Tkatchenko, A., and M{\"u}ller, K.-R.
\newblock Schnet: A continuous-filter convolutional neural network for modeling
  quantum interactions.
\newblock \emph{Advances in neural information processing systems}, 30, 2017.

\bibitem[Sifre \& Mallat(2014)Sifre and Mallat]{sifre2014rigid}
Sifre, L. and Mallat, S.
\newblock Rigid-motion scattering for texture classification.
\newblock \emph{arXiv preprint arXiv:1403.1687}, 2014.

\bibitem[Simonyan \& Zisserman(2014)Simonyan and Zisserman]{simonyan2014very}
Simonyan, K. and Zisserman, A.
\newblock Very deep convolutional networks for large-scale image recognition.
\newblock \emph{arXiv preprint arXiv:1409.1556}, 2014.

\bibitem[Sitzmann et~al.(2020)Sitzmann, Martel, Bergman, Lindell, and
  Wetzstein]{sitzmann2020implicit}
Sitzmann, V., Martel, J., Bergman, A., Lindell, D., and Wetzstein, G.
\newblock Implicit neural representations with periodic activation functions.
\newblock \emph{Advances in Neural Information Processing Systems},
  33:\penalty0 7462--7473, 2020.

\bibitem[Szegedy et~al.(2015)Szegedy, Liu, Jia, Sermanet, Reed, Anguelov,
  Erhan, Vanhoucke, and Rabinovich]{szegedy2015going}
Szegedy, C., Liu, W., Jia, Y., Sermanet, P., Reed, S., Anguelov, D., Erhan, D.,
  Vanhoucke, V., and Rabinovich, A.
\newblock Going deeper with convolutions.
\newblock In \emph{Proceedings of the IEEE conference on computer vision and
  pattern recognition}, pp.\  1--9, 2015.

\bibitem[Tan \& Le(2019)Tan and Le]{tan2019efficientnet}
Tan, M. and Le, Q.
\newblock Efficientnet: Rethinking model scaling for convolutional neural
  networks.
\newblock In \emph{International conference on machine learning}, pp.\
  6105--6114. PMLR, 2019.

\bibitem[Tay et~al.(2021)Tay, Dehghani, Abnar, Shen, Bahri, Pham, Rao, Yang,
  Ruder, and Metzler]{tay2021long}
Tay, Y., Dehghani, M., Abnar, S., Shen, Y., Bahri, D., Pham, P., Rao, J., Yang,
  L., Ruder, S., and Metzler, D.
\newblock Long range arena : A benchmark for efficient transformers.
\newblock In \emph{International Conference on Learning Representations}, 2021.
\newblock URL \url{https://openreview.net/forum?id=qVyeW-grC2k}.

\bibitem[Van Den~Oord et~al.(2016)Van Den~Oord, Dieleman, Zen, Simonyan,
  Vinyals, Graves, Kalchbrenner, Senior, and Kavukcuoglu]{van2016wavenet}
Van Den~Oord, A., Dieleman, S., Zen, H., Simonyan, K., Vinyals, O., Graves, A.,
  Kalchbrenner, N., Senior, A.~W., and Kavukcuoglu, K.
\newblock Wavenet: A generative model for raw audio.
\newblock \emph{SSW}, 125:\penalty0 2, 2016.

\bibitem[Warden(2018)]{warden2018speech}
Warden, P.
\newblock Speech commands: A dataset for limited-vocabulary speech recognition.
\newblock \emph{arXiv preprint arXiv:1804.03209}, 2018.

\bibitem[Wu et~al.(2019)Wu, Qi, and Fuxin]{wu2019pointconv}
Wu, W., Qi, Z., and Fuxin, L.
\newblock Pointconv: Deep convolutional networks on 3d point clouds.
\newblock In \emph{Proceedings of the IEEE/CVF Conference on Computer Vision
  and Pattern Recognition}, pp.\  9621--9630, 2019.

\bibitem[Xiong et~al.(2020)Xiong, Yang, He, Zheng, Zheng, Xing, Zhang, Lan,
  Wang, and Liu]{xiong2020layer}
Xiong, R., Yang, Y., He, D., Zheng, K., Zheng, S., Xing, C., Zhang, H., Lan,
  Y., Wang, L., and Liu, T.
\newblock On layer normalization in the transformer architecture.
\newblock In \emph{International Conference on Machine Learning}, pp.\
  10524--10533. PMLR, 2020.

\bibitem[Yadan(2019)]{Yadan2019Hydra}
Yadan, O.
\newblock Hydra - a framework for elegantly configuring complex applications.
\newblock Github, 2019.
\newblock URL \url{https://github.com/facebookresearch/hydra}.

\end{thebibliography}
\bibliographystyle{icml2022}

\newpage
\appendix
\onecolumn

\section*{{\LARGE Supplementary Material}\vspace{2mm} \\ {\Large Towards a General Purpose CNN for Long Range Dependencies in $\Nt$D}} \vspace{4mm}

\section{Dataset description}\label{appx:dataset_description}

\textbf{Sequential and Permuted MNIST.} The MNIST dataset \cite{lecun1998gradient} consists of 70{\sc{k}} gray-scale $28{\times}28$ handwritten digits divided into training validation and test sets of 60{\sc{k}} and 10{\sc{k}} samples, respectively. For validation purposes, the training dataset is further divided into training and validation sets of 55{\sc{k}} and 5{\sc{k}} samples, respectively. 

The sequential MNIST dataset (sMNIST) presents MNIST images as a sequence of 784 pixels for digit classification. Consequently, good predictions require the model to preserve long-term dependencies up to 784 steps in the past. The permuted MNIST dataset (pMNIST) incorporates an additional level of difficulty by permuting the order of all sMNIST sequences with a random permutation. Resultantly, models can no longer rely on local information for the construction of their features and the importance of modelling long-term dependencies becomes more pronounced.   

\textbf{CIFAR10, CIFAR100 and Sequential CIFAR10.} The CIFAR10 dataset \cite{krizhevsky2009learning} consists of 60{\sc{k}} real-world $32{\times}32$ RGB images uniformly drawn from 10 classes divided into training and test sets of 50{\sc{k}} and 10{\sc{k}} samples, respectively. The CIFAR100 dataset \cite{krizhevsky2009learning} is similar to the CIFAR10 dataset, with the difference that the images are now uniformly drawn from 100 different classes. For validation purposes, the training dataset of both CIFAR10 and CIFAR100 are further divided into training and validation sets of 45{\sc{k}} and 5{\sc{k}} samples, respectively. 

Analogously to the sMNIST dataset, the sequential CIFAR10 (sCIFAR10) dataset presents CIFAR10 images as a sequence of 1024 pixels for image classification. This dataset is more difficult than sMNIST, as \emph{(i)} larger memory horizons are required to successfully solve the task, and \emph{(ii)} more complex structures and intra-class variations are present in the images. 

\textbf{Speech Commands.} The Speech Commands dataset \cite{warden2018speech} consists of 105809 one-second audio recordings of 35 spoken words sampled at $16$kHz. Following \citet{kidger2020neural}, we extract 34975 recordings from ten spoken words to construct a balanced classification problem. We refer to this dataset as \textit{Raw Speech Commands}. In addition, we use the preprocessing steps of \citet{kidger2020neural} and extract mel-frequency cepstrum coefficients from the raw data. The resulting dataset, referred to as \textit{MFCC Speech Commands}, consists of time series of length 161 and 20 channels. 

\textbf{Long Range Arena.} The Long Range Arena benchmark \citep{tay2021long} consists of 6 tasks with lengths 1{\sc{k}}-16{\sc{k}} steps encompassing modalities and objectives that require similarity, structural, and visuospatial reasoning. The \texttt{Pathfinder}, \texttt{Path-X} and \texttt{Image} tasks are similar in nature to the sMNIST and sCIFAR10 tasks. These tasks consists of classification tasks performed on images that are treated as sequences.

\begin{wrapfigure}{r}{0.48\textwidth}
    \vspace{-4mm}
    \centering
    \hfill
         \begin{subfigure}[b]{0.23\textwidth}
         \centering
         \includegraphics[width=\textwidth]{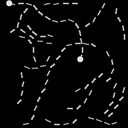}
         \caption{}
         \label{fig:positive}
     \end{subfigure}
     \hfill
         \begin{subfigure}[b]{0.23\textwidth}
         \centering
         \includegraphics[width=\textwidth]{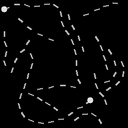}
         \caption{}
         \label{fig:negative}
     \end{subfigure}
    \vspace{-4mm}
    \caption{Positive and negative samples from the \texttt{Path-X} dataset
    \vspace{-5mm}}
    \label{fig:pathx_samples}
\end{wrapfigure}

The \texttt{Image} task corresponds to the sequential CIFAR10 dataset with the only difference that the CIFAR10 images are treated as gray-scale images. The \texttt{Pathfinder} and \texttt{Path-X} tasks are binary tasks in which binary images are provided and the model must predict whether the two points in the images are connected with a line or not --see Fig.~\ref{fig:pathx_samples} for an example--. The difference between both datasets is their resolution. Whereas \texttt{Pathfinder} has images of size 32$\times$32, \texttt{Path-X} has images of size 128$\times$128. It is important to mention that these tasks are so difficult that even if treated as 2D signals, CNNs without global receptive fields are unable to solve them \citep{gu2022efficiently}.

\textbf{STL-10.} The STL-10 dataset \citep{coates2011analysis} is a subset of the ImageNet dataset \citep{krizhevsky2012imagenet} consisting of 13,000 $96{\times}96$ real-world RGB images uniformly drawn from 10 classes divided into training
and test sets of 5{\sc{k}} and 8{\sc{k}} images, respectively. For validation purposes, the training dataset is further divided into training and validation sets of 4,500 and 500 samples, respectively. 

\section{Experimental details}\label{appx:empirical_details}
\vspace{-1mm}
\subsection{General remarks}
\vspace{-1mm}
\textbf{Code repository and logging.} Our code is written in \texttt{PyTorch}. We utilize \texttt{wandb} \citep{wandb} \texttt{hydra} \citep{Yadan2019Hydra} and \texttt{pytorch-lightning} \citep{falcon2019pytorch} for logging and code structuring. Our experiments are performed on NVIDIA TITAN RTX, A6000 and A100 GPUs, depending on the size of the datasets and inputs considered, and our code is publicly available at \href{https://github.com/david-knigge/ccnn}{\texttt{github.com/david-knigge/ccnn}}.

\textbf{The kernel generator network $\mathrm{G_{Kernel}}$.}
Our kernel generator network is parameterized as a 3-layer MAGNet \cite{romero2022flexconv} with 32 hidden units for the CCNN$_{4,140}$ models, and 64 hidden units for the larger CCNN$_{6, 380}$ models. The output size of the kernel generator network corresponds to the input channels of each layer in the network.

\textbf{Normalized relative positions.} The kernel generator network $\mathrm{G_{Kernel}}$ can, in principle, receive arbitrary coordinates as input. However, considering unitary step-wise relative positions, i.e., 0, 1, 2, ... , $\Nt$, can be problematic from a numerical stability perspective as $\Nt$ may grow very large, e.g., $\Nt{=}$16000 for the Speech Commands dataset. Consequently, based on insights from the Implicit Neural Representations, e.g., \citet{sitzmann2020implicit, fathony2021multiplicative}, we normalize the coordinates such that they lie in the space $[-1, 1]^{\Dt}$ for $\Dt$-dimensional kernels. To this end, we map largest unitary positions seen during training $[0, N]$ to a uniform linear space in $[-1, 1]$.
\vspace{-2mm}
\subsection{Hyperparameters and training details}
\vspace{-1mm}
\textbf{Optimizer and learning rate scheduler.} All our models are optimized with AdamW \citep{loshchilov2017decoupled} in combination with a cosine annealing learning rate scheduler \cite{loshchilov2016sgdr} and a linear learning rate warm-up stage of 10 epochs.

\textbf{Best hyperparameters found.} We perform hyperparameter search on the learning rate, dropout rate, weight decay, and $\omega_0$ of our CCNNs for each task considered.\footnote{$\omega_0$ serves as a prior on the variance of the data that is fitted with several types of implicit neural representations, e.g., SIRENs \cite{sitzmann2020implicit}, MFNs \cite{fathony2021multiplicative}, etc.} The best hyperparameters found are reported in Tables~\ref{tab:hyperparams_small} and \ref{tab:hyperparams_big}.

\begin{table}
    \centering
    \caption{Best hyperparameters found for CCNN$_{4, 140}$ on all tasks considered.}
    \label{tab:hyperparams_small}
    \begin{small}
    \scalebox{0.75}{
    \begin{tabular}{ccccccc}
    \toprule
         & $\boldsymbol{w_0}$ & \textbf{Dropout} & \textbf{Learning Rate} & \textbf{Weight Decay} & \textbf{Batch Size} & \textbf{Epochs}  \\
         \midrule
         \sc{sMNIST} & 2976.49 & 0.1 & 0.01 & 1e-6 & 100 & 210 \\
         \sc{pMNIST} & 2985.63 & 0.2 & 0.02 & 0 & 100 & 210 \\
         \sc{sCIFAR10} & 2386.49 & 0.0 & 0.02 & 0 & 50 & 210 \\
         \midrule
         \sc{Speech Commands (raw)} & 1295.61 & 0.2 & 0.02 & 1e-6 & 20 & 160 \\
         \sc{Speech Commands (mfcc)} & 750.18 & 0.2 & 0.02 & 1e-6 & 100 & 110 \\
         \midrule
         \sc{ListOps} & 784.66 & 0.1 & 0.001 & 1e-6 & 50 & 60\\
         \sc{Text} & 2966.60& 0.2 & 0.001 & 1e-5 & 50 & 60 \\
         \sc{Image} & 4005.15 & 0.2 & 0.01 & 0 & 50 & 210  \\
         \sc{Pathfinder} & 2272.56 & 0.2 & 0.01 & 0 & 100 & 210 \\
         \midrule 
         \sc{CIFAR10} & 1435.77 & 0.1 & 0.02 & 0.0001 & 50 & 210 \\
         \sc{CIFAR100} & 3521.55 & 0.1 & 0.02 & 0.0001 & 50 & 210 \\
         \sc{STL10} & 954.28 & 0.1 & 0.02 & 0 & 64 & 210 \\
         \sc{2DImage} & 2085.43& 0.2 & 0.02 & 1e-6 & 50 & 210  \\
         \sc{2DPathfinder}  & 1239.14 & 0.1 & 0.01 & 0 & 100 & 210 \\
         \bottomrule
    \end{tabular}}
    \end{small}
    \centering
    \caption{Best hyperparameters found for CCNN$_{6, 380}$ on all tasks considered.}
    \label{tab:hyperparams_big}
    \begin{small}
    \scalebox{0.75}{
    \begin{tabular}{ccccccc}
    \toprule
         & $\boldsymbol{w_0}$ & \textbf{Dropout} & \textbf{Learning Rate} & \textbf{Weight Decay} & \textbf{Batch Size} & \textbf{Epochs}  \\
         \midrule
         \sc{sMNIST} & 2976.49 & 0.1 & 0.01 & 0 & 100 & 210 \\
         \sc{pMNIST} & 2985.63 & 0.2 & 0.02 & 0 & 100 & 210 \\
         \sc{sCIFAR10} & 4005.15 & 0.25 & 0.01 & 0 & 50 & 210 \\
         \midrule
         \sc{Speech Commands (raw)} & 1295.61 & 0.2 & 0.02 & 1e-6 & 20 & 160 \\
         \sc{Speech Commands (mfcc)} & 750.18 & 0.2 & 0.02 & 1e-6 & 100 & 110 \\
         \midrule
         \sc{ListOps} & 784.66 & 0.25 & 0.001 & 0 & 50 & 60\\
         \sc{Text} & 2966.60& 0.3 & 0.02 & 0 & 50 & 60 \\
         \sc{Image} & 4005.15 & 0.1 & 0.01 & 0 & 50 & 210  \\
         \sc{Pathfinder} & 2272.56 & 0.1 & 0.01 & 1e-6 & 100 & 210 \\
         \midrule 
         \sc{CIFAR10} & 1435.77 & 0.15 & 0.02 & 0 & 50 & 210 \\
         \sc{CIFAR100} & 679.14 & 0.2 & 0.02 & 0 & 50 & 210 \\
         \sc{STL10} & 954.28 & 0.1 & 0.01 & 1e-6 & 64 & 210 \\
         \sc{2DImage} & 2306.08& 0.2 & 0.02 & 0 & 50 & 210  \\
         \sc{2DPathfinder}  & 3908.32 & 0.2 & 0.01 & 0 & 100 & 210 \\
         \bottomrule
    \end{tabular}}
    \end{small}
\end{table}


\end{document}